\documentclass[journal]{IEEEtran}
\usepackage[table]{xcolor}
\definecolor{mygray}{rgb}{0.9, 0.95, 1}

\usepackage{graphicx}
\usepackage{booktabs}
\usepackage{multirow}
\usepackage{array}
\usepackage{rotating}
\usepackage{algorithmic}
\usepackage{algorithm}
\usepackage{textcomp}
\usepackage{stfloats}
\usepackage{url}
\usepackage{verbatim}
\usepackage{pifont}
\usepackage[normalem]{ulem}
\useunder{\uline}{\ul}{}
\usepackage{times}
\usepackage{epsfig}
\usepackage{amsmath}
\usepackage{amssymb}
\usepackage{float}
\usepackage{wrapfig}
\usepackage{caption}
\usepackage{subcaption}
\usepackage{adjustbox}
\usepackage{cite}

\usepackage[accsupp]{axessibility} 
\providecommand{\eg}{e.g.}

\begin{document}

\title{VIGIL: Part-Grounded Structured Reasoning for Generalizable Deepfake Detection}

\author{Xinghan Li,
        Junhao Xu,
        and Jingjing Chen,~\IEEEmembership{Member,~IEEE}
\thanks{X. Li, J. Xu, and J. Chen are with College of Computer Science and Artificial Intelligence, Fudan University, Shanghai, China. (e-mail: \{xinghanli24, junhaoxu23\}@m.fudan.edu.cn, chenjingjing@fudan.edu.cn). (Corresponding author: Jingjing Chen.)}}


\maketitle

\begin{abstract}
Multimodal large language models (MLLMs) offer a promising path toward interpretable deepfake detection by generating textual explanations. However, the reasoning process of current MLLM-based methods combines evidence generation and manipulation localization into a unified step. This combination blurs the boundary between faithful observations and hallucinated explanations, leading to unreliable conclusions. Building on this, we present VIGIL, a part-centric structured forensic framework inspired by expert forensic practice through a plan-then-examine pipeline: the model first plans which facial parts warrant inspection based on global visual cues, then examines each part with independently sourced forensic evidence. A stage-gated injection mechanism delivers part-level forensic evidence only during examination, ensuring that part selection remains driven by the model's own perception rather than biased by external signals. We further propose a progressive three-stage training paradigm whose reinforcement learning stage employs part-aware rewards to enforce anatomical validity and evidence--conclusion coherence. To enable rigorous generalizability evaluation, we construct OmniFake, a hierarchical 5-Level benchmark where the model, trained on only three foundational generators, is progressively tested up to in-the-wild social-media data. Extensive experiments on OmniFake and cross-dataset evaluations demonstrate that VIGIL consistently outperforms both expert detectors and concurrent MLLM-based methods across all generalizability levels.
Project Page: \url{https://vigil.best}

\end{abstract}

\section{Introduction}
\label{sec:intro}

\IEEEPARstart{R}{ecent} advances~\cite{esser2024scaling,tian2024visual} in generative AI have greatly expanded digital content creation capabilities, yet simultaneously amplified the risk of deepfake misuse. This poses serious challenges to public trust and social media integrity. Deepfake detection, which aims to distinguish authentic facial images from synthetically generated ones, has consequently emerged as a critical research frontier. Although prior CNN-based expert models~\cite{CNNSpot, Fusing, LNP} and frequency-domain detectors~\cite{qian2020thinking, FreDect} have achieved strong in-distribution performance, they are built upon binary classification paradigms. While traditional localization techniques~\cite{GradCAM} can highlight suspicious regions, these methods cannot reason about \textit{why} those regions are suspicious or synthesize evidence across multiple regions into a coherent forensic conclusion.

To address this interpretability gap, recent studies incorporate multimodal large language models (MLLMs)~\cite{liu2023visual} into deepfake detection~\cite{zhang2024common, huang2024ffaa, Veritas}, seeking structured textual explanations through visual reasoning. While these efforts mark progress, three limitations remain underaddressed.
First, prior studies~\cite{jia2024can, IdentityAware} have noted that MLLM-based detectors tend to produce vague explanations. The textual claims (\eg, ``unnatural skin'', ``boundary artifacts'') lack spatial grounding and cannot be traced to concrete facial regions, making genuine analysis indistinguishable from hallucination.
Second, the global visual encoder of a general-purpose MLLM inherently suppresses high-frequency details critical for forensic analysis. Directly fine-tuning this encoder risks degrading the model's reasoning capabilities~\cite{IdentityAware}, while introducing additional specialized encoders~\cite{chen2025eagle, tong2024cambrian1} requires substantial data and computation for alignment. Although recent methods~\cite{peng2025mllm, FakeVLM} attempt to bridge this gap by injecting external forensic signals, such signals are delivered through global feature concatenation or image-level fusion, lacking structured correspondence to the specific regions referenced during reasoning.
Third, even when chain-of-thought reasoning is employed~\cite{Veritas}, the evidence cited in each step is generated by the model itself rather than drawn from an independent forensic source. The model acts as both claimant and evidence provider, with no external verification. These limitations share a common structural root, namely the lack of a mechanism that decouples claims from independently sourced evidence.
This leads to a natural question: \textit{can we restructure the reasoning process so that every claim is grounded in independently verifiable, region-specific forensic evidence?}

We answer this with \textbf{VIGIL}, a part-centric structured forensic framework. Our key observation is that forensic reasoning over facial images is naturally \textit{part-localized}, as forgery traces concentrate in anatomical regions and vary across generation methods. Inspired by human forensic reasoning, VIGIL first \textit{plans} which parts warrant examination based on global visual cues, then \textit{examines} each part with independently sourced region-specific evidence. This design enables \textit{reasoning reversion}, where accumulated part-level evidence can overturn initially erroneous judgments during inference.

Building upon this architecture, \textbf{Context-Aware Dynamic Signal Injection} bridges low-level physical evidence and high-level semantic reasoning by aggregating frequency-domain and pixel-level features into part-level evidence embeddings. A stage-gated mechanism restricts such evidence to the examination stage, preserving autonomous planning while supporting each examined region with dedicated forensic evidence.

Finally, to ensure genuine evidence reasoning rather than template memorization, we propose a progressive three-stage training paradigm: supervised fine-tuning on signal-semantic annotations, hard-sample self-training, and reinforcement learning with forensic-specific rewards for anatomical validity and evidence--conclusion coherence.

Moreover, existing evaluation protocols lack a systematic hierarchy of generalizability levels. We construct \textbf{OmniFake}, a 5-Level benchmark (Sec.~\ref{sec:omnifake}) where the model, trained on only three foundational generators, is progressively tested across increasingly challenging levels---from in-domain to cross-task localized manipulations and in-the-wild social-media data, covering the latest generators such as Nano Banana and Veo~3 (Fig.~\ref{fig:omnifake}). Extensive experiments demonstrate that VIGIL consistently outperforms both expert detectors and concurrent MLLM-based methods across all levels.

Our main contributions are:
\begin{itemize}
\item We propose \textbf{VIGIL}, a part-centric structured forensic framework where the model plans which facial parts to inspect, examines each with independently sourced forensic evidence, and synthesizes part-level findings into a final verdict, enabling verifiable, part-grounded explanations.

\item We introduce \textbf{context-aware dynamic signal injection} with a stage-gated mechanism that delivers part-level forensic evidence only during examination, preserving autonomous planning. We further design a progressive three-stage training pipeline with part-aware multi-dimensional rewards.

\item We construct \textbf{OmniFake}, a 5-Level benchmark with a hierarchical evaluation protocol designed for generalizability assessment, where the model trained solely on basic generators is progressively tested up to in-the-wild social media data.
\end{itemize}


\section{Related Work}
\label{sec:related}

\subsection{Deepfake Detection and Datasets}

Traditional deepfake detection methods primarily formulate the task as binary classification. Spatial-domain approaches capture manipulation artifacts~\cite{UnivFD, NPR, AIDE, xu2025deepfake}, frequency-domain methods mine spectral fingerprints~\cite{qian2020thinking, SAFE}, and data augmentation strategies~\cite{li2020face, shiohara2022detecting, D3, li2025revealing, yu2023augmented, yin2024improving} enhance generalization. While achieving strong in-distribution performance, these methods lack interpretability and struggle to generalize across unseen generators or heavily compressed media~\cite{DDA, Co-SPY}. Meanwhile, the commonly adopted protocol of training on FaceForensics++~\cite{FaceForensics++} and testing on limited benchmarks~\cite{DFDC, CelebA, WildRF} lacks a systematic hierarchy of generalizability levels, a gap our OmniFake benchmark (Sec.~\ref{sec:omnifake}) addresses.

\subsection{MLLMs for Deepfake Detection}

To overcome the interpretability limitations of black-box detectors, recent studies~\cite{xu2024fakeshield,ji2025towards,lin2025guard,chen2024x2} have integrated Multimodal Large Language Models (MLLMs)~\cite{liu2023visual, li2026spatialimaginer} into deepfake detection. Early methods such as DD-VQA~\cite{zhang2024common} and FFAA~\cite{huang2024ffaa} fine-tune MLLMs on image-text pairs to generate textual explanations alongside binary predictions, while M2F2-Det~\cite{guo2025rethinking} leverages CLIP-based features with an LLM as a plug-in interpreter. However, these approaches typically first produce an answer and then generate post-hoc explanations, limiting the depth of forensic reasoning. To improve the reliability of explanations, subsequent frameworks enforce structured reasoning and spatial alignment: Veritas~\cite{Veritas} introduces pattern-aware chains of thought to simulate human forensic processes, MARE~\cite{MARE} constructs text-spatially aligned preference data for multimodal reinforcement alignment, and TruthLens~\cite{TruthLens} explicitly aligns textual explanations with region-specific visual features. Recent advancements also explore Group Relative Policy Optimization (GRPO)~\cite{shao2024deepseekmath} to structurally align reasoning chains with visual evidence~\cite{Veritas, RLSBI}, and incorporate external low-level forensic signals to complement the MLLM's semantic encoding~\cite{peng2025mllm, FakeVLM}. Despite these advances, existing methods treat the entire image as a single target without grounding claims in specific regions, and the MLLM's global encoder suppresses high-frequency forensic details. VIGIL addresses both limitations through part-centric structured reasoning with context-aware forensic signal injection.


\section{OmniFake Dataset}
\label{sec:omnifake}

In this section, we introduce OmniFake, a hierarchical 5-Level benchmark with over 200K images designed for fine-grained generalizability evaluation in deepfake detection, as illustrated in Fig.~\ref{fig:omnifake}. Details are in Supplementary Material.

\begin{figure*}[t]
    \centering
    \includegraphics[width=\linewidth]{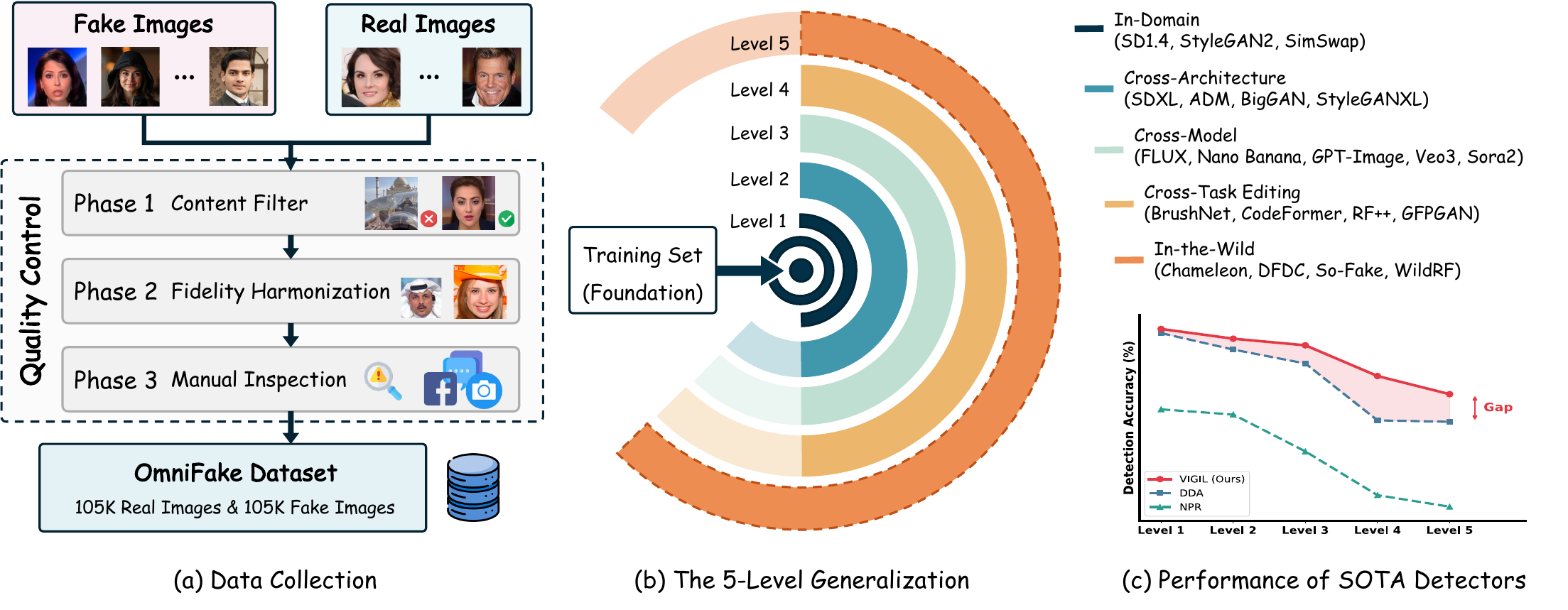}
    \caption{\textbf{Overview of the OmniFake dataset.} (a) Data collection and quality control pipeline. (b) The hierarchical 5-Level generalization protocol. The model is trained solely on foundational generators and progressively evaluated at each level with increasing difficulty. (c) Performance of existing detectors across levels, showing a widening generalization gap at higher levels.}
    \label{fig:omnifake}
\end{figure*}

\subsection{Data Collection}

\noindent \textbf{Data Sources.}
OmniFake is built from diverse real-face datasets and fake images spanning full-face synthesis, face swapping, localized editing, and in-the-wild social-media forgeries.
The training set is intentionally restricted to three foundational generators (StyleGAN2, SD~1.4, and SimSwap), while the test set covers increasingly distant generators, cross-task manipulations, and real-world degradations.

\noindent \textbf{Quality Control.}
We apply source-specific filtering, semantic validation for localized edits, manual verification for wild data, and face-identity de-duplication between training and testing splits.
After filtering and balancing, OmniFake contains approximately 105K real and 105K fake images.
Detailed source composition and quality-control protocols are provided in the supplementary material.

\subsection{Evaluation Protocol}
\textbf{Training.} The training set contains 40K images (20K real, 20K fake). We enforce a minimal prior knowledge setting: the model is only exposed to DF40 real images and fake images from three foundational generators (StyleGAN2, SD 1.4, SimSwap), covering basic GAN generation, early diffusion, and conventional face swap. Various generation paradigms and advanced models remain entirely unseen during training, simulating a practical forensic scenario.

\noindent \textbf{Evaluation.} The test protocol is organized into five levels of increasing difficulty, progressively evaluating generalization from familiar to entirely unseen scenarios:

\begin{itemize}
\item \textbf{Level 1: In-Distribution} (13K). Testing images share the same generation sources as training (StyleGAN2~\cite{stylegan2}, SD 1.4~\cite{sd1-4}, SimSwap~\cite{simswap}) but with entirely unseen identities, establishing a baseline performance metric.
\item \textbf{Level 2: Cross-Architecture} (24K). Fake images are generated by unseen models within related paradigm families, including ADM~\cite{ADM}, BigGAN~\cite{BigGAN}, SDXL~\cite{SDXL}, and StyleGANXL~\cite{StyleGANXL}. This level tests generalization \textit{within} known paradigms --- whether the model captures paradigm-level patterns (\eg, GAN upsampling artifacts) rather than memorizing model-specific fingerprints.
\item \textbf{Level 3: Cross-Model} (75K). Extends \textit{beyond} all paradigms seen during training to fundamentally different generation principles: face swap methods from FaceForensics++~\cite{FaceForensics++}, flow matching (FLUX~\cite{FLUX}, SD3~\cite{SD3}), autoregressive generation (NOVA~\cite{NOVA}), unified generation frameworks (Harmon~\cite{Harmon}), commercial generators (Midjourney~\cite{Midjourney}, Nano Banana~\cite{NanoBanana}, GPT-Image~\cite{GPT-Image}), and video generation models (Sora2~\cite{Sora2}, Veo 3~\cite{Veo3-1}). This level tests generalization to entirely unseen generation principles, where the model cannot rely on any paradigm-level prior from training.
\item \textbf{Level 4: Cross-Task} (15K). Shifts from fully synthesized faces to localized manipulations of real images, including inpainting (BrushNet~\cite{BrushNet}) and face restoration (CodeFormer~\cite{CodeFormer}, RestoreFormer++~\cite{RestoreFormer++}, GFPGAN~\cite{GFPGAN}). This level evaluates whether the model can detect subtle, localized edits rather than full-face synthesis.
\item \textbf{Level 5: In-the-Wild} (42K). Both real and fake images are sourced from social media and wild datasets (DFDC~\cite{DFDC}, Chameleon~\cite{AIDE}, WildRF~\cite{WildRF}, So-Fake~\cite{So-Fake}), featuring unknown generation methods compounded by real-world degradations such as compression and resizing.
\end{itemize}


\section{Method}
\label{sec:method}

\begin{figure*}[t]
    \centering
    \includegraphics[width=\linewidth]{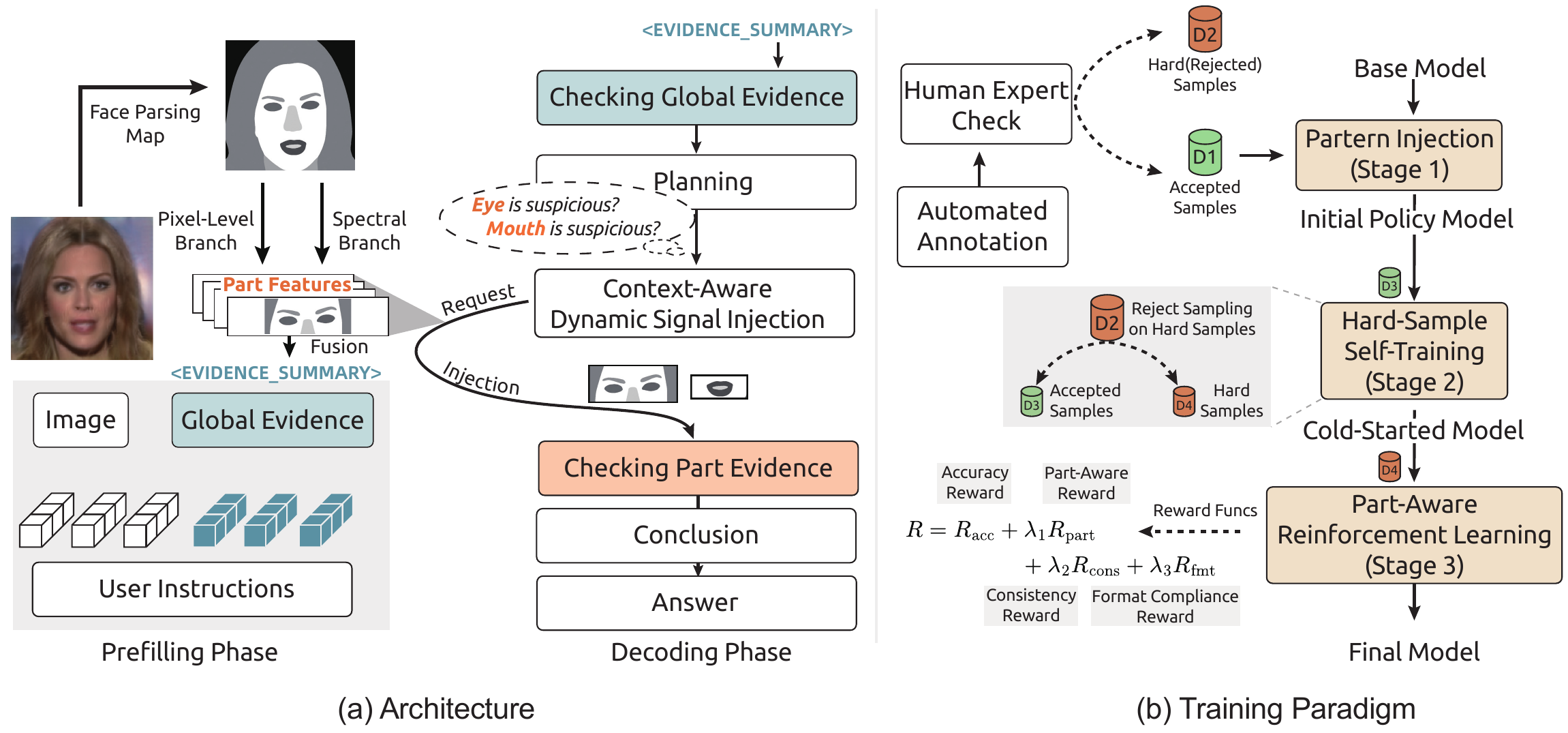}
    \caption{\textbf{Overview of VIGIL.}
\textit{Left}: the part-centric forensic architecture. Specialized forensic encoders extract frequency-domain and pixel-level features, aggregated into part-level evidence embeddings via face parsing masks. A global evidence summary is injected before reasoning; part-level evidence is delivered only during examination through stage-gated injection.
\textit{Right}: the progressive three-stage training paradigm. Stage~1 performs supervised fine-tuning on signal-semantic annotations; Stage~2 expands hard samples via rejection sampling; Stage~3 applies part-aware reinforcement learning.}
    \label{fig:method}
\end{figure*}

In this section, we present VIGIL in detail. We first describe the part-centric forensic architecture (Sec.~\ref{sec:injection}), and then the progressive training paradigm that transforms this architecture into a capable forensic reasoner (Sec.~\ref{sec:training}). An overview is shown in Fig.~\ref{fig:method}.

\subsection{Part-Centric Forensic Architecture}
\label{sec:injection}

To decouple claim generation from evidence sourcing, VIGIL organizes reasoning as a plan-then-examine forensic pipeline with five stages: \textit{global evidence}, \textit{planning} (specifying semantic parts to inspect), \textit{part-level evidence} (per-part examination), \textit{conclusion}, and \textit{answer}. The key challenge is bridging the gap between low-level physical signals and high-level structured reasoning. We address this through context-aware dynamic signal injection that extracts forensic signals from specialized encoders, aggregates them at the part level, and injects them on demand into the MLLM's reasoning process.

\textbf{Low-Level Forensic Feature Extraction.}
The pre-trained visual backbone of a general-purpose MLLM is optimized for semantic understanding, and tends to overlook fine-grained physical traces critical for forensic analysis. To provide the examination stage with low-level evidence, we introduce two complementary forensic encoders:
(1)~A \textit{spectral branch} applies FFT-based spectral analysis with a bank of high-pass filters, followed by a convolutional backbone, to extract frequency-domain anomaly features.
(2)~A \textit{pixel-level branch} leverages a pre-trained DINOv3~\cite{DINOv3} to extract fine-grained visual features sensitive to pixel-level inconsistencies.
Given an input image $\mathbf{I}$, we obtain frequency feature maps $\mathbf{F}^{\text{freq}} \!\in\! \mathbb{R}^{C_f \times H \times W}$ with a spatial anomaly attention map $\mathbf{A}^{\text{freq}} \!\in\! \mathbb{R}^{H \times W}$, and pixel-level feature maps $\mathbf{F}^{\text{pixel}} \!\in\! \mathbb{R}^{C_p \times H \times W}$.


\textbf{Part-Level Evidence Aggregation.}
We leverage face semantic parsing~\cite{FaceParser} to define a fixed set of $K\!=\!8$ anatomical regions (\eg, eyes, nose, mouth, skin), each with a binary mask $\mathbf{M}_k$.
For each part $k$, mask-guided average pooling extracts region-specific features from both branches, which are concatenated and projected into a part evidence embedding:
\begin{equation}
    \begin{aligned}
    \mathbf{e}_k ={}& \mathrm{MLP}\!\left([\,\mathrm{AvgPool}(\mathbf{F}^{\text{freq}}, \mathbf{M}_k)\;;\right.\\
    &\left.\mathrm{AvgPool}(\mathbf{F}^{\text{pixel}}, \mathbf{M}_k)\,]\right) \in \mathbb{R}^D,
    \end{aligned}
    \label{eq:part_embed}
\end{equation}
where $D$ is the evidence embedding dimension. Parts with empty masks are assigned a learnable default vector.
To form a global evidence summary, we aggregate part embeddings using a soft attention mechanism informed by frequency-domain responses. A per-part score $a_k \!=\! \frac{1}{|\mathbf{M}_k|}\sum_{(h,w) \in \mathbf{M}_k} \mathbf{A}^{\text{freq}}_{h,w}$ serves as an attention prior:
\begin{equation}
    \mathbf{e}_g = \sum_{k=1}^{K} w_k \, \mathbf{e}_k, \quad w_k = \frac{\exp(a_k)}{\sum_{j=1}^{K}\exp(a_j)},
    \label{eq:global_embed}
\end{equation}
This soft weighting emphasizes regions with frequency-domain irregularities, while preserving a holistic evidence summary for later reasoning.

\textbf{Context-Aware Dynamic Signal Injection.}
We design a dual-path injection mechanism to deliver forensic evidence into the MLLM's token embedding space. We register a dedicated \texttt{<EVIDENCE\_SUMMARY>} token and $K\!=\!8$ anatomical part tokens (\eg, \texttt{<P\_LEFT\_EYE>}, \texttt{<P\_NOSE>}, \texttt{<P\_MOUTH>}, covering eyes, eyebrows, nose, mouth, face contour, and hair) as special tokenizer tokens. Let $\mathbf{h}$ denote an original token embedding. The \textit{global path} injects $\mathbf{e}_g$ into the \texttt{<EVIDENCE\_SUMMARY>} token, providing an overall forensic prior before reasoning begins. The \textit{contextual part path} injects $\mathbf{e}_k$ into corresponding part tokens during the examination stage. Both paths follow the same residual pattern:
\begin{equation}
    \hat{\mathbf{h}}_{\text{global}} = \mathbf{h}_{\text{global}} + \alpha \cdot \mathbf{e}_g, \qquad
    \hat{\mathbf{h}}_k = \mathbf{h}_k + \gamma \cdot \mathbf{e}_k,
    \label{eq:injection}
\end{equation}
where $\alpha$ and $\gamma$ are zero-initialized learnable scalars.

The contextual part path further employs a stage-gated injection mechanism: part-level evidence (Eq.~\eqref{eq:injection}, right) is injected only when a part token appears within the \texttt{<part\_evidence>} block during autoregressive generation; when the same token appears in the \texttt{<planning>} block, no part evidence is injected ($\hat{\mathbf{h}}_k \!=\! \mathbf{h}_k$). This ensures that part selection during planning is driven by the model's own visual assessment, not by external signal strength.

\textbf{Semantic Binding of Part Tokens.}
Each part token is bound to mask-pooled features of its corresponding anatomical region (Eq.~\ref{eq:part_embed}), so the examination stage receives region-specific visual content rather than a textual trigger. Additional implementation details are in the supplementary material.

\subsection{Progressive Training Paradigm}
\label{sec:training}

The plan-then-examine pipeline requires structured part-level outputs, forensic grounding, and evidence--conclusion coherence. We therefore adopt a three-stage paradigm: Stage~1 establishes signal--semantics alignment through supervised fine-tuning, Stage~2 broadens coverage to hard cases via rejection sampling, and Stage~3 refines reasoning quality through reinforcement learning with part-aware rewards.

\begin{figure*}[t]
    \centering
    \includegraphics[width=\linewidth]{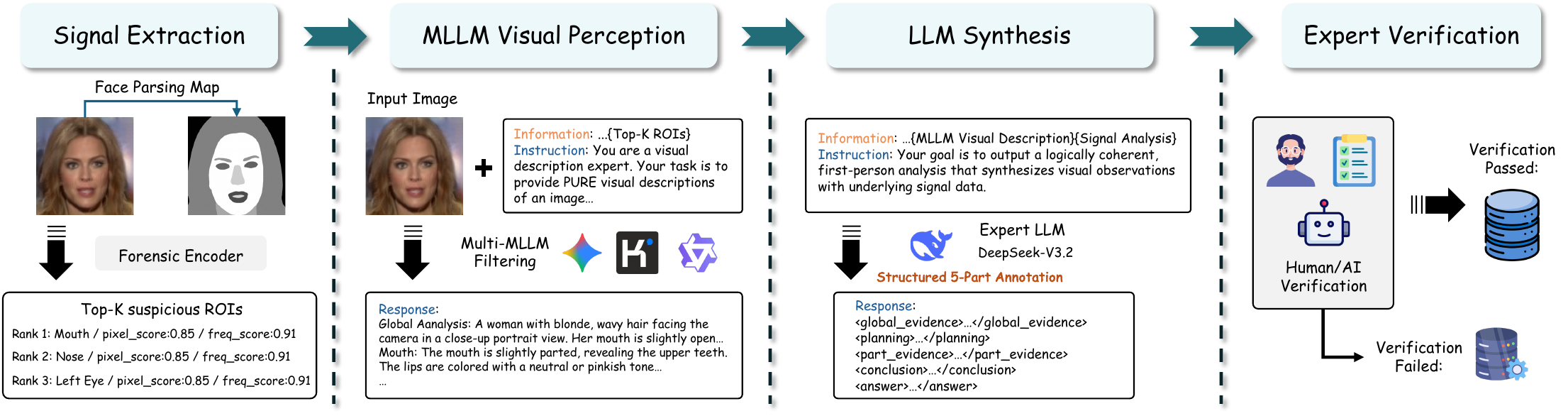}
    \caption{\textbf{Signal-semantic annotation pipeline.}
Given an input image, forensic encoders extract frequency anomaly maps and pixel-level features, aggregated into part-level anomaly scores to identify suspicious regions (Step~1). Multiple off-the-shelf MLLMs produce visual descriptions, with consensus filtering to remove hallucinated observations (Step~2). Finally, an LLM expert synthesizes signal analysis and visual descriptions into structured five-part annotations (Step~3).}
    \label{fig:annotation}
\end{figure*}

\textbf{Stage 1: Signal-Semantic Annotation and Structured Reasoning SFT.}
\label{sec:annotation}
We design an automated annotation pipeline that integrates low-level forensic analysis with high-level semantic understanding, as illustrated in Fig.~\ref{fig:annotation}.
It first scores part-level anomalies from forensic signals, then obtains consensus visual descriptions from multiple off-the-shelf MLLMs, and finally uses an LLM expert to synthesize both sources into structured five-part annotations.
Generated annotations are verified by domain experts; detailed prompts and filtering rules are provided in the supplementary material.

We denote the successfully annotated samples as $\mathcal{D}_1$ and the samples for which annotation fails (\eg, the pipeline cannot produce internally consistent reasoning) as $\mathcal{D}_2$. Stage~1 performs supervised fine-tuning on $\mathcal{D}_1 \!=\! \{(\mathbf{q}, \mathbf{s})_i\}_{i=1}^{N_1}$ with a standard autoregressive objective.

\textbf{Stage 2: Hard-Sample Self-Training.}
Rather than discarding $\mathcal{D}_2$, we leverage the Stage-1 model---now an initial policy with basic forensic ability---to mine correct reasoning trajectories through rejection sampling.
For each sample in $\mathcal{D}_2$ with ground-truth label $y$, we generate $M$ candidate responses via temperature sampling and retain those simultaneously satisfying structural validity and prediction correctness.
The retained trajectories are further verified by domain experts. We denote the curated set as $\mathcal{D}_3$ and fine-tune the Stage-1 model on $\mathcal{D}_3$.
Samples in $\mathcal{D}_2$ that remain unsolved after rejection sampling form $\mathcal{D}_4$, which is reserved for Stage~3.

\textbf{Stage 3: Part-Aware Reinforcement Learning.}
To further refine reasoning quality on the remaining cases $\mathcal{D}_4$, we employ Group Relative Policy Optimization (GRPO)~\cite{shao2024deepseekmath} with multi-dimensional forensic rewards.
For each query $\mathbf{q} \in \mathcal{D}_4$ with binary label $y$, GRPO samples $G$ responses from the current policy and optimizes using group-normalized advantages with a KL penalty against the Stage-2 reference policy $\pi_{\text{ref}}$.

The reward $R_i$ for each response is evaluated from four dimensions. Beyond standard accuracy reward $R_{\text{acc}}$ and format reward $R_{\text{fmt}}$, we introduce two specific rewards:

\textit{Part-Aware Reward} ($R_{\text{part}}$) evaluates the quality of part-level reasoning from three perspectives:
(i)~\textit{Plan-evidence consistency}: the $F_1$ score between the part sets mentioned in the Planning and Examine stages, ensuring the model examines exactly what it planned.
(ii)~\textit{Spatial existence verification}: the face parsing model~\cite{FaceParser} checks whether each planned part actually exists in the input image, penalizing fabricated regions.
(iii)~\textit{Quantity constraint}: a penalty for listing excessive parts, encouraging focus on the most suspicious regions.

\textit{Consistency Reward} ($R_{\text{cons}}$) prevents fabricated reasoning where the model decides a conclusion first and then invents supporting evidence.
We feed the model's evidence (global evidence, planning, and part evidence, \textit{excluding} the final answer) to an independent judge LLM $\mathcal{J}$, which predicts a label solely from the evidence:
\begin{equation}
    R_{\text{cons}} = \mathbb{I}\!\left(\mathcal{J}(\text{evidence}) = \hat{y}\right) \cdot \mathbb{I}(\hat{y} = y),
\end{equation}
where $\hat{y}$ is the model's predicted label and $y$ is the ground truth. The reward is granted only when the judge's independent assessment aligns with the model's conclusion and the conclusion is correct.

The final reward aggregates all dimensions:
\begin{equation}
    R = R_{\text{acc}} + \lambda_1 R_{\text{part}} + \lambda_2 R_{\text{cons}} + \lambda_3 R_{\text{fmt}},
\end{equation}

where $\lambda_1\!=\!0.4$, $\lambda_2\!=\!0.4$, $\lambda_3\!=\!0.2$ are balancing coefficients.
Detailed reward definitions are in the supplementary material.


\section{Experiment}
\label{sec:experiment}

\subsection{Experimental Setup}

\textbf{State-of-the-Art Methods.} We compare VIGIL with three groups of baselines. The first group contains eight expert binary detectors: UnivFD~\cite{UnivFD}, NPR~\cite{NPR}, SAFE~\cite{SAFE}, AIDE~\cite{AIDE}, C2P-CLIP~\cite{C2P-CLIP}, D$^3$~\cite{D3}, Co-SPY~\cite{Co-SPY}, and DDA~\cite{DDA}. These detectors are retrained for three epochs on each benchmark split. The second group contains five general-purpose MLLMs evaluated zero-shot with task-specific prompts: Qwen3-VL-8B~\cite{Qwen3-VL}, GLM-4.6V~\cite{glm-4.6v}, Kimi-K2.5~\cite{kimi-k2.5}, GPT-5.5~\cite{gpt-5.5}, and Gemini-3-Pro~\cite{gemini-3-pro}. The third group contains MLLM-based deepfake detectors, Veritas~\cite{Veritas} and FakeVLM~\cite{FakeVLM}. On OmniFake, we reconstruct each method's training data with its published annotation pipeline using the same categories of foundation models as ours, namely MLLMs and LLMs. On HydraFake~\cite{Veritas}, we report published Veritas and FakeVLM results, and train VIGIL on the HydraFake training split with our annotation pipeline.

\textbf{Metrics.} Following established protocols in deepfake detection~\cite{Veritas}, we use accuracy as the primary metric. Precision and recall in the supplementary material show consistent trends.

\textbf{Implementation Details.} VIGIL uses Qwen3-VL-8B~\cite{Qwen3-VL} with full fine-tuning at all stages. Stage~1 trains for three epochs with a learning rate of $5\!\times\!10^{-5}$ and batch size 1. Stage~2 uses the same setting for one epoch. Stage~3 uses a learning rate of $1\!\times\!10^{-6}$, batch size 8, eight rollouts, and temperature 1.0. DeepSeek-V3.2~\cite{Deepseek-v3.2} serves as the judge LLM $\mathcal{J}$. Each stage initializes from the previous checkpoint; additional details are in the supplementary material.

\subsection{State-of-the-art Comparison}


\begin{table*}[t]
    \centering
    \caption{\textbf{Performance comparison on OmniFake dataset.} In-domain (ID) results are averaged. The best results are \textbf{bolded} and second best are \underline{underlined}.}
    \label{tab:omnifake}
    \begin{adjustbox}{width=\linewidth}
    \renewcommand{\arraystretch}{1.25}
      \begin{tabular}{l c cccc cccccccccc cccc cccc c}
        \toprule
        \multirow{2}{*}{\textbf{Method}} & \multirow{2}{*}{\textbf{ID}} & \multicolumn{4}{c}{\textbf{Cross-Arch.}} & \multicolumn{10}{c}{\textbf{Cross-Model}} & \multicolumn{4}{c}{\textbf{Cross-Task}} & \multicolumn{4}{c}{\textbf{In-the-Wild}} & \multirow{2}{*}{\textbf{Avg.}} \\
        \cmidrule(lr){3-6} \cmidrule(lr){7-16} \cmidrule(lr){17-20} \cmidrule(lr){21-24}
        & & \rotatebox{60}{ADM} & \rotatebox{60}{BigGAN} & \rotatebox{60}{SDXL} & \rotatebox{60}{StyleGANXL} & \rotatebox{60}{FF++} & \rotatebox{60}{FLUX} & \rotatebox{60}{GPT-Image} & \rotatebox{60}{Harmon} & \rotatebox{60}{Midjourney} & \rotatebox{60}{NOVA} & \rotatebox{60}{Nano} & \rotatebox{60}{SD3} & \rotatebox{60}{Sora2} & \rotatebox{60}{Veo3} & \rotatebox{60}{BrushNet} & \rotatebox{60}{CodeF.} & \rotatebox{60}{GFPGAN} & \rotatebox{60}{RestoreF.} & \rotatebox{60}{Chameleon} & \rotatebox{60}{DFDC} & \rotatebox{60}{So-Fake} & \rotatebox{60}{WildRF} & \\
        \midrule
        UnivFD (\textit{CVPR'23}) & 77.4 & 72.6 & 92.0 & 91.3 & 88.0 & 59.7 & 88.8 & 95.1 & 66.3 & 83.0 & 70.7 & 79.0 & \underline{97.0} & 74.6 & 72.0 & 62.1 & 64.1 & 58.3 & 66.8 & 76.0 & 53.1 & 88.3 & 85.0 & 76.6 \\
        NPR (\textit{CVPR'24}) & 83.1 & 68.9 & 87.9 & 86.7 & 84.9 & 56.3 & 84.5 & 91.1 & 62.9 & 78.9 & 67.0 & 74.9 & 95.5 & 71.2 & 67.6 & 59.0 & 68.1 & 64.6 & 74.5 & 70.6 & 49.2 & 74.2 & 63.3 & 73.3 \\
        SAFE (\textit{KDD'25}) & 79.4 & 70.1 & 74.4 & 73.9 & 74.9 & 56.4 & 72.0 & 74.3 & 65.2 & 78.8 & 54.8 & 59.9 & 74.4 & 59.5 & 50.4 & 59.8 & 74.5 & 62.1 & 75.5 & 63.9 & 40.1 & 73.3 & 66.3 & 66.7 \\
        AIDE (\textit{ICLR'25}) & 72.3 & 92.1 & 91.0 & 90.7 & 82.4 & 65.3 & 93.8 & 93.5 & 91.1 & 83.9 & 77.9 & 71.3 & 94.2 & 71.4 & 73.5 & 76.0 & 66.9 & 62.8 & 64.5 & 75.5 & 47.3 & 90.6 & 86.4 & 78.9 \\
        C2P-CLIP (\textit{AAAI'25}) & 79.0 & 71.1 & 72.2 & 73.8 & 80.8 & 51.1 & 84.1 & 73.8 & 63.4 & 83.1 & 72.8 & 72.9 & 83.9 & 73.6 & 63.4 & 70.4 & 68.8 & 65.2 & 68.8 & 72.7 & 50.2 & 82.5 & 82.0 & 72.2 \\
        D$^3$ (\textit{CVPR'25}) & 82.7 & 89.4 & 97.7 & 92.0 & 91.3 & 64.5 & 96.7 & 96.2 & 86.3 & 88.3 & 90.0 & 87.8 & 96.7 & 62.4 & 73.8 & 62.1 & 60.6 & 62.2 & 57.5 & 82.7 & 61.3 & 86.7 & 84.8 & 80.6 \\
        Co-SPY (\textit{CVPR'25}) & 83.5 & 89.9 & 91.3 & 95.9 & 82.4 & 65.0 & 96.5 & 96.4 & 94.5 & 95.1 & 92.7 & \underline{94.0} & 96.2 & 71.5 & \underline{88.4} & 75.5 & 71.4 & 70.1 & 68.3 & 84.9 & 61.9 & \textbf{95.8} & \underline{87.4} & 84.7 \\
        DDA (\textit{NeurIPS'25}) & \underline{97.8} & 95.1 & 93.8 & 95.6 & \textbf{94.0} & \underline{80.9} & 97.1 & \underline{97.1} & \underline{97.3} & \underline{95.4} & \underline{93.8} & 93.6 & 96.8 & 80.7 & 86.7 & 79.3 & \underline{83.5} & 81.9 & 79.2 & 82.3 & \underline{70.3} & 85.9 & 84.3 & 88.8 \\
        \midrule
        Qwen3-VL-8B & 54.4 & 67.8 & 82.3 & 61.2 & 50.7 & 53.6 & 60.6 & 62.4 & 84.9 & 52.2 & 58.2 & 51.6 & 47.3 & 59.7 & 48.9 & 61.2 & 48.4 & 51.2 & 53.5 & 57.9 & 65.1 & 57.1 & 76.7 & 59.4 \\
        GLM-4.6V & 54.4 & 62.4 & 72.0 & 52.5 & 54.1 & 53.7 & 53.0 & 74.3 & 72.1 & 54.4 & 64.4 & 55.4 & 53.0 & 64.8 & 54.1 & 65.0 & 49.7 & 50.4 & 52.0 & 60.6 & 54.9 & 59.0 & 76.5 & 59.2 \\
        Kimi-K2.5 & 65.1 & 67.3 & 82.8 & 79.1 & 66.3 & 67.9 & 80.7 & 85.1 & 87.3 & 61.8 & 79.1 & 62.8 & 77.5 & 62.2 & 65.2 & 72.1 & 53.4 & 57.5 & 61.8 & 76.7 & 58.9 & 70.5 & 74.1 & 70.2 \\
        GPT-5.5 & 66.7 & 78.5 & 90.6 & 85.6 & 78.9 & 60.4 & 86.1 & 83.9 & 93.0 & 63.8 & 76.3 & 65.9 & 80.5 & 69.2 & 62.8 & 68.4 & 57.6 & 56.2 & 53.7 & 73.6 & 56.9 & 72.9 & 77.0 & 72.1 \\
        Gemini-3-Pro & 72.3 & 80.7 & 92.1 & 89.3 & 81.5 & 62.3 & 88.7 & 85.6 & 91.6 & 72.3 & 91.6 & 66.8 & 79.0 & 78.9 & 68.2 & 72.9 & 56.8 & 56.4 & 57.8 & 69.3 & 52.7 & 83.6 & 73.8 & 75.0 \\
        \midrule
        FakeVLM (\textit{NeurIPS'25}) & 83.5 & 77.5 & 85.0 & 82.0 & 79.3 & 69.4 & 81.5 & 80.0 & 77.5 & 81.5 & 76.8 & 75.0 & 79.0 & 74.8 & 72.8 & 74.8 & 76.5 & 74.2 & 72.3 & 75.3 & 67.8 & 80.8 & 75.5 & 77.1 \\
        Veritas (\textit{ICLR'26}) & 96.8 & 92.8 & \underline{97.8} & 96.5 & 92.5 & 74.2 & \underline{97.8} & 96.5 & 89.5 & 94.3 & 93.0 & 87.5 & 95.8 & 86.5 & 84.0 & 80.5 & 83.2 & 78.3 & 73.8 & 83.5 & 67.3 & 90.2 & 83.5 & 87.6 \\
        \midrule
        \textit{\textbf{VIGIL Cold-Start}} & 97.8 & \underline{95.5} & 96.1 & \textbf{99.6} & 90.3 & 80.3 & 94.1 & 94.9 & 96.2 & 93.7 & 92.7 & 91.2 & 95.4 & \underline{87.5} & 85.4 & \underline{86.5} & 83.3 & \underline{84.2} & \textbf{85.6} & \underline{87.5} & 67.1 & \underline{95.0} & 85.0 & \underline{89.8} \\
        \rowcolor{mygray}
        \textit{\textbf{VIGIL (Ours)}} & \textbf{98.6} & \textbf{95.9} & \textbf{98.8} & \underline{99.4} & \underline{92.8} & \textbf{82.1} & \textbf{97.9} & \textbf{98.1} & \textbf{99.4} & \textbf{96.1} & \textbf{96.5} & \textbf{95.2} & \textbf{98.8} & \textbf{98.4} & \textbf{92.0} & \textbf{96.4} & \textbf{91.1} & \textbf{88.9} & \underline{81.7} & \textbf{88.0} & \textbf{72.2} & 92.4 & \textbf{91.5} & \textbf{93.1} \\
        \bottomrule
      \end{tabular}
    \end{adjustbox}
\end{table*}


\begin{table*}[t]
    \centering
    \caption{\textbf{Performance comparison (Acc\%) on HydraFake dataset.} In-domain (ID) results are averaged. The best results are \textbf{bolded} and second best are \underline{underlined}.}
    \label{tab:hydrafake}
    \begin{adjustbox}{width=\linewidth}
    \renewcommand{\arraystretch}{1.25}
      \begin{tabular}{l c cccccc cccccc cccccc c}
        \toprule
        \multirow{2}{*}{\textbf{Method}} & \multirow{2}{*}{\textbf{ID}} & \multicolumn{6}{c}{\textbf{Cross-Model}} & \multicolumn{6}{c}{\textbf{Cross-Forgery}} & \multicolumn{6}{c}{\textbf{Cross-Domain}} & \multirow{2}{*}{\textbf{Avg.}} \\
        \cmidrule(lr){3-8} \cmidrule(lr){9-14} \cmidrule(lr){15-20}
        & & \rotatebox{60}{ADF} & \rotatebox{60}{FLUX} & \rotatebox{60}{StarryAI} & \rotatebox{60}{MAGI-1} & \rotatebox{60}{HART} & \rotatebox{60}{Infinity} & \rotatebox{60}{St.GAN2} & \rotatebox{60}{ICLight} & \rotatebox{60}{CodeF.} & \rotatebox{60}{InfiniteY.} & \rotatebox{60}{PuLID} & \rotatebox{60}{FaceAda.} & \rotatebox{60}{Deepface.} & \rotatebox{60}{InfiniteY.} & \rotatebox{60}{Dreamina} & \rotatebox{60}{HailuoAI} & \rotatebox{60}{GPT-4o} & \rotatebox{60}{FFIW} & \\
        \midrule
        UnivFD (\textit{CVPR'23}) & 83.2 & 89.0 & 93.0 & 81.6 & 73.7 & 95.1 & 92.1 & 60.8 & 81.6 & 74.0 & 72.9 & 68.1 & 79.8 & 66.6 & 67.7 & 80.6 & 74.4 & 73.6 & 68.3 & 77.7 \\
        NPR (\textit{CVPR'24}) & 86.4 & 65.2 & 95.0 & 62.3 & 97.0 & 85.3 & 83.2 & 48.3 & 52.3 & 52.1 & 89.5 & 87.5 & 66.7 & 55.5 & 69.7 & 81.6 & 59.0 & 53.7 & 54.2 & 70.8 \\
        SAFE (\textit{KDD'25}) & 73.4 & 63.3 & 73.8 & 69.8 & 72.3 & \underline{99.4} & 63.6 & 49.9 & \textbf{97.4} & 50.5 & 50.2 & 74.1 & 53.4 & 59.9 & 51.4 & 83.5 & 54.9 & 50.1 & 49.7 & 65.3 \\
        AIDE (\textit{ICLR'25}) & 77.3 & 84.0 & 81.3 & 88.0 & 81.3 & 85.1 & 64.7 & 54.4 & 73.9 & 86.1 & 77.8 & 59.9 & 78.2 & 54.3 & 59.7 & 64.9 & 58.0 & 50.8 & 51.9 & 70.1 \\
        C2P-CLIP (\textit{AAAI'25}) & 87.3 & 66.2 & 75.5 & 85.0 & 79.4 & 82.2 & 80.5 & 65.0 & 67.1 & 81.9 & 74.6 & 76.7 & 88.2 & 59.7 & 70.3 & 72.2 & 69.6 & 51.8 & 69.6 & 73.8 \\
        D$^3$ (\textit{CVPR'25}) & 93.4 & \textbf{96.7} & 98.3 & 93.3 & 97.1 & 99.2 & 98.2 & 66.3 & 73.2 & 93.6 & 80.2 & 81.0 & 90.1 & \textbf{78.9} & 71.4 & 92.2 & 86.2 & \underline{85.7} & 70.2 & 86.6 \\
        Co-SPY (\textit{CVPR'25}) & 88.9 & 89.8 & 94.2 & 82.8 & 92.1 & 94.5 & 93.3 & 81.0 & 87.7 & 90.1 & 89.1 & 76.1 & \underline{94.2} & 73.3 & 84.2 & 72.6 & 79.2 & 80.8 & 66.2 & 84.7 \\
        DDA (\textit{NeurIPS'25}) & 95.3 & 81.7 & 97.2 & 95.3 & 96.6 & 97.0 & 96.5 & \underline{94.1} & \underline{91.8} & 96.8 & \underline{94.7} & 82.9 & \textbf{95.9} & 63.9 & \underline{85.8} & 64.9 & 67.1 & 52.4 & \textbf{79.3} & 85.7 \\
        \midrule
        Qwen3-VL-8B & 64.8 & 57.6 & 56.3 & 51.2 & 49.3 & 80.0 & 62.8 & 53.6 & 53.9 & 55.0 & 53.9 & 57.1 & 55.9 & 51.8 & 56.9 & 69.6 & 71.2 & 8.8 & 54.3 & 56.0 \\
        GLM-4.6V & 60.0 & 55.0 & 50.1 & 51.5 & 50.5 & 57.4 & 51.4 & 54.1 & 49.5 & 53.7 & 52.9 & 55.7 & 50.1 & 53.7 & 71.7 & 59.6 & 75.5 & 18.7 & 51.1 & 53.8 \\
        Kimi-K2.5 & 64.3 & 91.2 & 81.9 & 64.5 & 56.0 & 88.5 & 77.0 & 60.4 & 64.6 & 68.6 & 60.1 & 61.8 & 71.2 & 62.9 & 75.9 & \textbf{95.2} & 78.0 & 60.9 & 53.3 & 70.3 \\
        GPT-5.5 & 72.7 & 91.1 & 83.8 & 67.2 & 59.5 & 85.5 & 75.8 & 78.2 & 67.3 & 65.2 & 76.8 & 66.2 & 71.8 & 62.5 & 74.3 & 52.3 & 82.5 & 26.6 & 54.1 & 69.1 \\
        Gemini-3-Pro & 70.0 & 87.5 & 87.8 & 75.9 & 49.7 & 84.0 & 87.2 & 76.3 & 63.0 & 69.9 & 79.6 & 80.3 & 75.3 & 62.8 & 78.3 & 85.5 & 79.3 & 41.8 & 53.1 & 73.0 \\
        \midrule
        FakeVLM (\textit{NeurIPS'25}) & 78.5 & 78.2 & 78.5 & 77.0 & 74.5 & 76.5 & 76.8 & 70.8 & 76.2 & 76.2 & 76.9 & 76.5 & 77.7 & \underline{75.7} & 83.6 & 81.5 & 80.8 & 78.7 & 74.5 & 77.3 \\
        Veritas (\textit{ICLR'26}) & \underline{97.3} & 94.8 & \textbf{99.8} & \textbf{97.0} & \textbf{99.9} & \textbf{99.9} & \textbf{99.9} & 90.3 & 75.7 & \underline{97.0} & 91.8 & \underline{95.1} & 91.7 & 58.6 & 84.1 & 92.3 & \underline{90.2} & \textbf{89.2} & \underline{78.5} & \underline{90.7} \\
        \midrule
        \rowcolor{mygray}
        \textit{\textbf{VIGIL (Ours)}} & \textbf{98.2} & \underline{96.3} & \underline{99.5} & \underline{96.2} & \underline{97.2} & 99.1 & \underline{99.6} & \textbf{95.5} & 88.0 & \textbf{97.1} & \textbf{94.9} & \textbf{95.9} & 93.4 & 67.5 & \textbf{86.9} & \underline{94.1} & \textbf{91.5} & 85.7 & 73.5 & \textbf{92.1} \\
        \bottomrule
      \end{tabular}
    \end{adjustbox}
\end{table*}

\textbf{Results on OmniFake.}
As shown in Table~\ref{tab:omnifake}, VIGIL achieves 93.1\% overall accuracy, outperforming DDA by 4.3 points and Veritas by 5.5 points.
Existing expert detectors remain competitive on cross-architecture and cross-model splits, but degrade on localized edits and in-the-wild data. General-purpose MLLMs are also weak; even Gemini-3-Pro reaches only 75.0\%, and the Qwen3-VL-8B backbone achieves 59.4\% zero-shot.
VIGIL's gains are largest on harder settings: it reaches 89.5\% on Level~4 cross-task data, exceeding Veritas by 10.5 points and DDA by 8.5 points, and maintains 86.0\% on Level~5 in-the-wild data.
These results support the value of part-level examination for localized manipulations and degraded wild images, where holistic cues are less reliable. The full model improves over Cold-Start by 3.3 points, mainly on L3--L5, indicating that progressive training improves out-of-distribution generalization.

\noindent  \textbf{Results on HydraFake.}
As shown in Table~\ref{tab:hydrafake}, VIGIL achieves 92.1\% overall accuracy on HydraFake.
The most notable gain over Veritas appears in cross-forgery scenarios, where VIGIL improves the averaged accuracy by 3.8 points. On ICLight, VIGIL reaches 88.0\%, compared with 75.7\% for Veritas, further confirming the effectiveness of part-centric reasoning on unseen forgery types.

\subsection{Ablation Studies}

\begin{figure*}[t]
    \scriptsize
    \centering
    \begin{minipage}{0.3\linewidth}
        \captionof{table}{\textbf{Core contribution ablation.}}
        \label{tab:ablation_arch}
        \renewcommand{\arraystretch}{1.15}
        \setlength{\tabcolsep}{2.5pt}
        \centering
        \begin{adjustbox}{max width=\linewidth}
        \begin{tabular}{l cccccc}
            \toprule
            \textbf{Variant} & \textbf{L1} & \textbf{L2} & \textbf{L3} & \textbf{L4} & \textbf{L5} & \textbf{Avg$^\dagger$} \\
            \midrule
            w/o Both & 91.0 & 85.1 & 82.8 & 78.5 & 73.5 & 82.2 \\
            w/o Part-Centric & 94.2 & 90.3 & 87.3 & 80.5 & 78.0 & 86.1 \\
            w/o Forensic & 93.2 & 89.2 & 88.1 & 80.1 & 74.5 & 85.0 \\
            \midrule
            \rowcolor{mygray}
            \textbf{VIGIL (Cold-Start)} & \textbf{97.8} & \textbf{95.4} & \textbf{91.1} & \textbf{84.9} & \textbf{83.7} & \textbf{90.6} \\
            \bottomrule
        \end{tabular}
        \end{adjustbox}
    \end{minipage}
    \hfill
    \begin{minipage}{0.34\linewidth}
        \captionof{table}{\textbf{Training paradigm ablation. S1--S3 denote the three training stages.}}
        \label{tab:ablation_train}
        \renewcommand{\arraystretch}{1.2}
        \setlength{\tabcolsep}{4.0pt}
        \centering
        \scalebox{1.0}{
        \begin{tabular}{ccc cccccc}
            \toprule
            \textbf{S1} & \textbf{S2} & \textbf{S3} & \textbf{L1} & \textbf{L2} & \textbf{L3} & \textbf{L4} & \textbf{L5} & \textbf{Avg$^\dagger$} \\
            \midrule
            \checkmark & & & 96.6 & 95.5 & 90.2 & 81.7 & 78.2 & 88.4 \\
            \checkmark & \checkmark & & 97.8 & 95.4 & 91.1 & 84.9 & 83.7 & 90.6 \\
            \checkmark & & \checkmark & 97.2 & 96.9 & 93.3 & 86.9 & 85.7 & 92.0 \\
            \rowcolor{mygray}
            \checkmark & \checkmark & \checkmark & \textbf{98.6} & \textbf{96.7} & \textbf{95.5} & \textbf{89.5} & \textbf{86.0} & \textbf{93.3} \\
            \bottomrule
        \end{tabular}
        }
    \end{minipage}
    \hfill
    \begin{minipage}{0.281\linewidth}
        \captionof{table}{\textbf{Ablation on the reward functions in Training Stage 3.}}
        \label{tab:ablation_reward}
        \renewcommand{\arraystretch}{1.0}
        \setlength{\tabcolsep}{7.0pt}
        \centering
        \begin{adjustbox}{max width=\linewidth}
        \begin{tabular}{cccc c}
            \toprule
            $R_{\text{acc}}$ & $R_{\text{fmt}}$ & $R_{\text{part}}$ & $R_{\text{cons}}$ & \textbf{Avg$^\dagger$} \\
            \midrule
            \checkmark & & & & 90.6 \\
            \checkmark & \checkmark & & & 90.9 \\
            \checkmark & \checkmark & \checkmark & & 91.8 \\
            \checkmark & \checkmark & & \checkmark & 91.9 \\
            \rowcolor{mygray}
            \checkmark & \checkmark & \checkmark & \checkmark & \textbf{93.3} \\
            \bottomrule
        \end{tabular}
        \end{adjustbox}
    \end{minipage}
\end{figure*}


\begin{figure*}[t]
    \scriptsize
    \centering
    \begin{minipage}{0.46\linewidth}
        \captionof{table}{\textbf{Evaluation of reasoning quality.} We utilize score and pairwise ELO rating.}
        \label{tab:reason_quality}
        \renewcommand{\arraystretch}{1.1}
        \setlength{\tabcolsep}{10.0pt}
        \centering
        \scalebox{0.92}{
        \begin{tabular}{l cc c}
            \toprule
            \multirow{2}{*}{\textbf{Model}} & \multicolumn{2}{c}{\textbf{Score Evaluation}} & \multirow{2}{*}{\textbf{ELO Rating}} \\
            \cmidrule(lr){2-3}
            & GPT-5.5 & Gemini-3-Pro & \\
            \midrule
            Qwen3-VL-8B   & 2.2301 & 2.3563 & 511.7 \\
            Gemini-3-Pro      & 3.4614 & 3.0847 & 747.2 \\
            Veritas   & 4.1705 & 4.0418 & 989.1 \\
            \midrule
            \textit{VIGIL} (Cold-Start) & 4.2635 & 4.1380 & 1026.4 \\
            \textit{VIGIL} (w/ $R_\text{acc}$+$R_\text{fmt}$) & 4.6319 & 4.3042 & 1234.7 \\
            \rowcolor{mygray}
            \textit{\textbf{VIGIL (Ours)}} & \textbf{4.7446} & \textbf{4.4853} & \textbf{1389.3} \\
            \bottomrule
        \end{tabular}
        }
    \end{minipage}
    \hfill
    \begin{minipage}{0.50\linewidth}
        \captionof{table}{\textbf{Robustness evaluation under Compression and blur on OmniFake.}}
        \label{tab:robustness}
        \renewcommand{\arraystretch}{1.1}
        \centering
        \scalebox{0.92}{
        \begin{tabular}{l c ccc ccc}
            \toprule
            \multirow{2}{*}{\textbf{Method}} & \multirow{2}{*}{\textbf{Orig.}} & \multicolumn{3}{c}{\textbf{JPEG Compression}} & \multicolumn{3}{c}{\textbf{Gaussian Blur}} \\
            \cmidrule(lr){3-5} \cmidrule(lr){6-8}
            & & QF=90 & QF=70 & QF=60 & $\sigma\!=\!1$ & $\sigma\!=\!2$ & $\sigma\!=\!4$ \\
            \midrule
            AIDE     & 78.9 & 78.4 & 76.2 & 75.1 & 77.3 & 75.2 & 71.0 \\
            C2P-CLIP & 72.2 & 52.8 & 52.4 & 52.0 & 51.3 & 50.2 & 49.9 \\
            D$^3$    & 80.6 & 80.3 & 78.5 & 77.4 & 79.5 & 77.9 & 74.6 \\
            Co-SPY   & 84.7 & 83.9 & 82.3 & 81.2 & 79.4 & 76.3 & 69.9 \\
            DDA      & 88.8 & 88.5 & 86.3 & 85.2 & 88.0 & 85.1 & 77.8 \\
            \midrule
            \rowcolor{mygray}
            \textit{\textbf{VIGIL (Ours)}} & 93.1 & 92.9 & 91.4 & 90.3 & 92.7 & 90.6 & 87.4 \\
            \bottomrule
        \end{tabular}
        }
    \end{minipage}
\end{figure*}

\begin{figure*}[t]
    \centering
    \includegraphics[width=\linewidth]{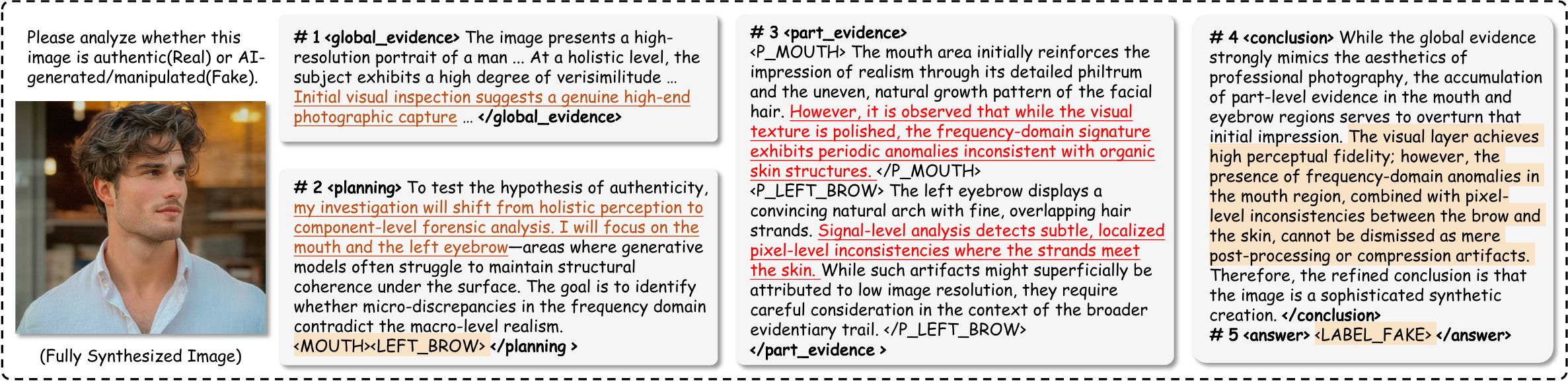}
    \caption{\textbf{Reasoning reversion case.} The model initially leans toward ``authentic'' based on global appearance, but reverses its judgment after examining part-level forensic evidence. The key transition point is highlighted.}
    \label{fig:reversion}
\end{figure*}

\textbf{Ablation Variant Definitions.}
Table~\ref{tab:ablation_arch} evaluates the Cold-Start model after Stages~1--2, isolating architectural effects from reinforcement learning.
The variants remove both core components, remove part-level decomposition while keeping global signal injection, or replace forensic embeddings with zero vectors.
Avg$^\dagger$ denotes the mean of five level-wise averages.
Additional signal-design ablations for stage gating, spectral features, and pixel-level features are provided in the supplementary material.

\textbf{Architecture Ablation.}
Removing both part-centric reasoning and forensic signals reduces average accuracy by 8.4 points.
Forensic signals and part-centric reasoning contribute 3.9 and 2.8 points individually, while their combined gain reaches 8.4 points, showing complementarity.
Forensic signals provide low-level evidence, while part-centric reasoning structures how the model uses that evidence. The gain is larger on harder levels, reaching 10.2 points on Level~5, where grounded regional evidence becomes more important as the domain gap widens.
Supplementary ablations show that the spectral branch is the most influential signal component and that stage-gated injection contributes 2.2 points overall, especially on harder levels where premature part-level signals may bias planning.

\textbf{Training and Reward Ablation.}
Tables~\ref{tab:ablation_train} and \ref{tab:ablation_reward} show that adding hard-sample self-training before GRPO raises accuracy from 92.0\% to 93.3\%.
Stage~2 expands coverage of challenging cases and provides better rollouts for reinforcement learning. For rewards, $R_{\text{part}}$ and $R_{\text{cons}}$ improve over the accuracy-plus-format baseline by 0.9 and 1.0 points, and combining all rewards yields the best result, with gains mainly on L3--L5.

\subsection{Further Analyses}

\textbf{Reasoning Quality Evaluation.}
Following~\cite{chen2024mllm,Veritas,zhou2025aigi}, we evaluate reasoning quality on 1K samples using judge scores and pairwise ELO.
Table~\ref{tab:reason_quality} shows that VIGIL achieves the best scores, with the full RL stage improving ELO by 354.5 points over Cold-Start.
The forensic-specific rewards add 176.3 ELO points beyond the accuracy-plus-format baseline, indicating that part-aware and consistency rewards improve not only prediction accuracy but also explanation quality.

\textbf{Robustness Evaluation.}
Table~\ref{tab:robustness} shows that VIGIL remains robust under JPEG compression and Gaussian blur, achieving 90.3\% at QF=60 and 87.4\% at $\sigma\!=\!4$ without corresponding data augmentation.
By contrast, DDA drops from 88.8\% to 77.8\% under the strongest blur.

\textbf{Qualitative Analysis: Reasoning Reversion.}
Fig.~\ref{fig:reversion} illustrates reasoning reversion. VIGIL initially leans toward ``authentic'' from global appearance, but part-level evidence from the mouth and left eyebrow reveals frequency and pixel-level inconsistencies. The accumulated evidence overturns the initial impression and leads to the correct forgery conclusion.

\section{Conclusion}
\label{sec:conclusion}

We present VIGIL, a part-centric forensic framework that decouples claim generation from independently sourced evidence through plan-then-examine reasoning and stage-gated evidence injection. Together with part-aware reinforcement learning, VIGIL grounds region-level explanations in forensic signals rather than template-like rationales.

Experiments on OmniFake and HydraFake show consistent improvements over expert detectors and MLLM-based methods, especially under cross-task and in-the-wild generalization. These results suggest that region-level evidence grounding is a promising direction for explainable visual forensics.

\bibliographystyle{IEEEtran}
\bibliography{main}

\vfill

\end{document}